\documentclass[10pt,twocolumn,letterpaper]{article}

\usepackage{cvpr}              

\usepackage{microtype}
\usepackage{multirow}
\usepackage{amsfonts}
\usepackage{bm}
\usepackage{tikz}
\usetikzlibrary{arrows.meta,decorations.pathreplacing,calc}
\usepackage[algo2e,ruled,vlined,linesnumbered]{algorithm2e}
\SetArgSty{textup}
\usepackage{xpatch}
\makeatletter
\xpatchcmd{\algocf@algorithm}{\if@twocolumn\@latex@error{[H] in two columns mode is not allowed for algorithms}\fi}{}{}{\PackageWarning{main}{Could not patch the [H] check of algorithm2e}}
\makeatother
\usepackage{xurl}
\usepackage{placeins}
\setlist{topsep=2pt, itemsep=2pt, parsep=0pt, partopsep=0pt}
\setlist[description]{leftmargin=1em, labelindent=0pt}
\setlist[itemize,enumerate]{leftmargin=*}

\definecolor{cvprblue}{rgb}{0.21,0.49,0.74}
\usepackage[pagebackref,breaklinks,colorlinks,allcolors=cvprblue]{hyperref}

\def\confName{CVPR}
\def\confYear{2027}

\newcommand{\vz}{\bm{z}}
\renewcommand{\vs}{\bm{s}}
\newcommand{\vx}{\bm{x}}
\newcommand{\vr}{\bm{r}}
\newcommand{\va}{\bm{a}}
\newcommand{\vy}{\bm{y}}
\newcommand{\vo}{\bm{o}}
\newcommand{\net}{f_\theta}
\newcommand{\rms}{\operatorname{RMSNorm}}
\newcommand{\amin}{a_{\min}}
\newcommand{\amax}{a_{\max}}

\newcommand{\strm}{STRM\xspace}
\newcommand{\ours}{\strm}

\title{Simpler Tiny Recursive Models}

\author{Torben Glass$^{*}$ \quad Zhengxi Zhang$^{*}$ \quad Paul Swoboda\\
Heinrich Heine University D\"usseldorf, Germany\\
{\tt\small \{togla104,zhengxi.zhang,paul.swoboda\}@hhu.de}
}

\begin{document}
\maketitle

\begin{NoHyper}
  \def\thefootnote{*}\footnotetext{Equal contribution.}
  \def\thefootnote{\arabic{footnote}}
\end{NoHyper}

\begin{abstract}
Small models that execute deep looped computational graphs have shown great experimental results on puzzle-like problems like Sudoku, finding shortest paths through mazes, etc.
The most well-known one is TRM (tiny recursive model), a simplification of HRM (hierarchical reasoning model).
However, the underlying recursion is complicated and involves a high-level and a low-level state, a nested loop recursion, and the network calls with different input and output signatures depending on where we are in the loops.
These mechanisms are needed to avoid memorization and overfitting.
We propose \ours, a simplified design with a single state and a uniform update.
\ours does not sacrifice generalization, improves training speed and final performance.
Our key architectural choices that allow our simpler recursion to work are exponential moving updates on the states with varying rates that are jittered during training and readout of the solution on slow-updating channels only.
We test \ours on Sudoku-Extreme, PPBench and Maze-Hard with the same data, parameters and compute as TRM: we improve over TRM by $3.8$ points on Sudoku-Extreme and $8.4$ points on PPBench and match it on Maze-Hard.
On ARC-AGI-1 and ARC-AGI-2, \ours improves over the published TRM results by $5.4$ and $0.5$ points.
We will release our code upon acceptance of the paper.
\end{abstract}

\section{Introduction}
\label{sec:intro}

Recently, small looping models have attracted great attention since they allow to attack problems which need much compute but few parameters~\citep{dehghani2019ut,gao2025urm,prairie2026parcae}.
However, simply looping small models is not working and a few mechanisms have to be adopted to make the recursion train stably, keep backprop in the large computational graph manageable and regularize so that overfitting does not lead to a large generalization gap.
One of the most elegant approaches for small puzzle-like problems is the tiny recursive model approach~\citep{jolicoeur2025trm}, borne out of earlier work on hierarchical reasoning models~\citep{wang2025hrm}.

The Hierarchical Reasoning Model (HRM)~\citep{wang2025hrm} splits the recursion into a fast and a slow recurrent module, inspired by hierarchical processing in the brain.
The slow module is updated only after the fast one has run for several steps, and only the slow state is decoded into the answer.
This hierarchy delays and funnels the flow of information, since intermediate results reach the output only through a slowly changing state, which we argue also regularizes the network.
To keep the long computational graph tractable, HRM backpropagates only through the final steps of the recursion and supervises it repeatedly along the way.

The Tiny Recursive Model (TRM)~\citep{jolicoeur2025trm} strips this design down to a single small network that alternately refines a latent state and an answer state, without the biological motivation and with simpler gradient and halting schemes.
This simplification also improves accuracy considerably, e.g.\ from $55.0\%$ to $87.4\%$ on Sudoku-Extreme.

Yet, we argue, we can still go further simplifying this recursive scheme.
First, we show how we can get rid of two states and keep only one.
Accompanying that, we also reduce the two different updates arranged in nested loops to a single one.
Doing this naively has been done in the TRM paper, but has not been satisfactory: The author reports on Sudoku-Extreme a performance drop of $87.4\%$ to $71.9\%$.
To stabilize training and add regularization we use an exponentially averaged state with log-spaced rates.
During training, we jitter the rates additionally.
Readout is performed on slow channels only.
These three simple tricks make our scheme work and match or improve results throughout all experiments we have performed.

In summary, our contributions are:
\begin{description}
    \item[Simplicity:] We propose a simplified, single state scheme for recursive models.
    \item[Stability:] To make training stable and prevent overfitting we propose exponential moving averages on the state with log-spaced rates jittered during training and readout on slow channels only.
    \item[Experiments:] With the same data, parameters and compute, \ours improves over TRM at the best checkpoints by $3.8$ points on Sudoku-Extreme and $8.4$ points on PPBench and matches it on Maze-Hard. Unlike TRM for Sudoku, \ours does not collapse late in training. On ARC-AGI-1 and ARC-AGI-2, it improves over the published TRM results by $5.4$ and $0.5$ points.
\end{description}

\section{Related work}

\label{sec:related}

\paragraph{Recursive reasoning with tiny networks.}
The Hierarchical Reasoning Model (HRM)~\citep{wang2025hrm} solves hard constraint-satisfaction puzzles with a small recurrent model trained from about one thousand examples, without pre-training or chain-of-thought.
It couples a slow high-level and a fast low-level recurrent module, motivated by multi-timescale processing in the brain, and trains them with a one-step gradient approximation, deep supervision and adaptive halting.
HRM also introduced Sudoku-Extreme, the benchmark we use, on which it reaches $55.0\%$.
The Tiny Recursive Model (TRM)~\citep{jolicoeur2025trm} simplifies HRM to a single two-layer network that alternately updates a latent state $\vz$ and an answer state $\vy$.
It backpropagates through one full recursion instead of relying on a fixed-point gradient and reaches $87.4\%$ on Sudoku-Extreme with $5$M parameters.
Both build on weight-shared recurrent depth as in Universal Transformers~\citep{dehghani2019ut}, on adaptive computation time~\citep{graves2016act} and PonderNet~\citep{banino2021pondernet}, and on the implicit-depth view of deep equilibrium models~\citep{bai2019deq}, from which HRM's one-step gradient derives.

\paragraph{Is the hierarchy needed?}
Several analyses question whether the two-level state of HRM and TRM matters: a model with only the low-level module performs about as well as HRM~\citep{ge2025hrmmisconceptions}, HRM's ARC-AGI performance stems mainly from the outer refinement loop~\citep{arcprize2025hrmanalysis}, single-state recurrent Transformers are comparable to HRM~\citep{rodrigues2026dissecting}, and the Universal Reasoning Model reaches strong ARC-AGI results with a plain looped Transformer~\citep{gao2025urm}.
TRM's own ablations, however, point the other way: a single state loses $15.5$ points on Sudoku-Extreme (\cref{sec:intro}), and a ``multi-scale'' variant with more latent features drops the accuracy from $87.4\%$ to $77.6\%$~\citep{jolicoeur2025trm}.
In language modelling, \citet{han2026hierflat} likewise report a gap between a two-speed shared-weight model and a flat Universal Transformer.
The closest prior work, the Simple Recursive Model (SRM)~\citep{liao2026srm}, conjectures that $\vy$ helps by giving access to earlier steps and replaces it by a skip connection that adds a context $f(\vz_0)$, computed once per supervision step, to every inner update of a single state.
Experimental results are very scarce, only Sudoku was tested and no proper number was given, however from a convergence plot the result looks roughly on par.
We share the single-state premise but replace this skip connection by a parameter-free bank of log-spaced exponential moving averages whose slow part alone is decoded, and evaluate on Sudoku-Extreme, six puzzle types of PPBench, Maze-Hard and ARC-AGI.

\begin{figure*}[t]
\centering
\definecolor{cgZ}{RGB}{38,104,170}
\definecolor{cgY}{RGB}{206,112,32}
\definecolor{cgSfast}{RGB}{214,236,229}
\definecolor{cgSslow}{RGB}{22,104,88}
\definecolor{cgGrad}{RGB}{176,48,62}
\tikzset{
  cg/net/.style={draw=#1, fill=#1!10, line width=0.6pt, rounded corners=3pt,
    minimum width=1.7cm, minimum height=0.62cm, inner sep=1.5pt, font=\scriptsize},
  cg/flow/.style={-{Latex[length=1.7mm, width=1.4mm]}, line width=0.9pt},
  cg/tap/.style={-{Latex[length=1.4mm, width=1.1mm]}, line width=0.6pt},
  cg/dot/.style={circle, fill=#1, inner sep=0pt, minimum size=3.2pt},
  cg/nograd/.style={fill=black!5, draw=black!22, rounded corners=5pt},
  cg/grad/.style={fill=cgGrad!5, draw=cgGrad!65, dashed, line width=0.6pt, rounded corners=5pt},
  cg/inner/.style={draw=cgZ!75, fill=cgZ!6, line width=0.8pt, rounded corners=4pt},
  cg/idx/.style={font=\tiny, text=cgZ!80!black, inner sep=1pt},
  cg/badge/.style={circle, fill=white, inner sep=0.5pt, minimum size=0.5cm, font=\scriptsize, text=#1},
  cg/mix/.style={circle, draw=cgSslow, fill=white, line width=0.7pt, inner sep=0pt, minimum size=0.3cm,
    font=\scriptsize, text=cgSslow},
  cg/tag/.style={font=\scriptsize, text=black!60, inner sep=1pt},
  cg/gradtag/.style={cg/tag, text=cgGrad!80!black},
  cg/brace/.style={decorate, decoration={brace, amplitude=4pt, raise=1pt}, line width=0.5pt, black!60},
  cg/title/.style={anchor=west, font=\small\bfseries},
  cg/subtitle/.style={anchor=west, font=\scriptsize, text=black!65},
  cg/state/.style={anchor=east, font=\small},
  cg/statenote/.style={anchor=east, font=\tiny, text=black!55},
  cg/out/.style={draw=#1, fill=#1!10, rounded corners=3pt, font=\scriptsize, align=center,
    inner sep=2.5pt, line width=0.6pt},
}
\newcommand{\cgloop}{\tikz[baseline=-0.55ex]\draw[-{Latex[length=0.9mm, width=0.9mm]}, line width=0.55pt, color=.]
  (60:0.11cm) arc[start angle=60, end angle=-250, radius=0.11cm];}
\begin{tikzpicture}[x=2.05cm, y=1cm]
  \begin{scope}
    \node[cg/title] at (-0.32, 3.05) {(a) TRM\quad\normalfont\scriptsize\color{black!65}two learned states $\vy$ and $\vz$, two update rules, two nested loops};
    \draw[cg/brace] (0.55, 2.35) -- node[above=4pt, cg/tag, text=black!75] {\cgloop\ \textbf{outer loop}: $H$ cycles; one segment $= H(n+1)$ network calls} (6.45, 2.35);
    \draw[cg/nograd] (0.55, -0.85) rectangle (3.45, 2.2);
    \draw[cg/grad] (3.55, -0.85) rectangle (6.45, 2.2);
    \node[cg/tag, anchor=north west] at (0.6, 2.17) {cycle $h = 1,\dots,H{-}1$\,: no gradient};
    \node[cg/gradtag, anchor=north west] at (3.6, 2.17) {cycle $h = H$\,: backpropagated};
    \foreach \a in {1, 4} {
      \draw[cg/inner] (\a - 0.45, -0.42) rectangle (\a + 1.45, 0.42);
      \node[cg/tag, text=cgZ!80!black, anchor=north] at (\a + 0.5, -0.44) {\cgloop\ \textbf{inner loop}: $n$ updates of $\vz$};
    }
    \node[cg/state, text=cgY!85!black] at (0.12, 1.3) {$\vy$};
    \node[cg/statenote] at (0.12, 0.98) {answer};
    \node[cg/state, text=cgZ!85!black] at (0.12, 0) {$\vz$};
    \node[cg/statenote] at (0.12, -0.32) {latent};
    \foreach \t in {1, 2, 4, 5} {
      \node[cg/net=cgZ] (z\t) at (\t, 0) {$\net(\vz,\, \vy + \vx)$};
      \node[cg/dot=cgY] (yd\t) at (\t, 1.3) {};
      \draw[cg/tap, cgY] (yd\t) -- (z\t.north);
    }
    \foreach \t in {3, 6} {
      \node[cg/net=cgY] (y\t) at (\t, 1.3) {$\net(\vy,\, \vz)$};
      \node[cg/dot=cgZ] (zd\t) at (\t, 0) {};
      \draw[cg/tap, cgZ] (zd\t) -- (y\t.south);
    }
    \draw[cg/flow, cgZ] (0.2, 0) -- (z1);
    \draw[cg/flow, cgZ] (z1) -- (z2);
    \draw[cg/flow, cgZ] (z2) -- (z4);
    \draw[cg/flow, cgZ] (z4) -- (z5);
    \draw[cg/flow, cgZ] (z5) -- (6.75, 0);
    \draw[cg/flow, cgY] (0.2, 1.3) -- (y3);
    \draw[cg/flow, cgY] (y3) -- (y6);
    \draw[cg/flow, cgY] (y6) -- (6.75, 1.3);
    \node[cg/out=cgY, anchor=west] at (6.78, 1.3) {heads\\[-1pt]$\hat{\bm{y}},\, q_{\text{halt}}$};
    \node[cg/tag, anchor=west] at (6.78, 0) {next segment};
  \end{scope}

  \begin{scope}[yshift=-4.1cm]
    \node[cg/title] at (-0.32, 2.4) {(b) \ours\quad\normalfont\scriptsize\color{black!65}one state, one update rule, one loop};
    \draw[cg/brace] (0.55, 1.7) -- node[above=4pt, cg/tag, text=black!75] {\cgloop\ \textbf{one loop}: $K = H(n+1)$ network calls} (6.45, 1.7);
    \draw[cg/nograd] (0.55, -0.55) rectangle (3.45, 1.55);
    \draw[cg/grad] (3.55, -0.55) rectangle (6.45, 1.55);
    \node[cg/tag, anchor=north west] at (0.6, 1.52) {calls $1,\dots,K{-}G$: no gradient};
    \node[cg/gradtag, anchor=north west] at (3.6, 1.52) {last $G = n{+}1$ calls: backpropagated};
    \node[cg/state, text=cgSslow] at (0.12, 0.3) {$\vs$};
    \node[cg/statenote] at (0.12, -0.02) {state};
    \foreach \t in {1, ..., 6} {
      \node[cg/dot=cgSslow] (d\t) at (\t - 0.55, 0.3) {};
      \node[cg/net=cgZ, minimum width=0.95cm] (f\t) at (\t - 0.13, 0.3) {$\net$};
      \node[cg/mix] (m\t) at (\t + 0.3, 0.3) {$+$};
      \draw[cg/flow, cgZ] (f\t) -- node[below=1pt, font=\tiny, text=cgZ!80!black, inner sep=1pt] {$\va$} (m\t);
      \draw[cg/tap, cgSslow, rounded corners=4pt] (d\t) |- (\t - 0.13, 0.82) -| (m\t.north);
    }
    \draw[cg/flow, cgSslow] (0.2, 0.3) -- (f1);
    \foreach \t [evaluate=\t as \n using int(\t + 1)] in {1, ..., 5} {
      \draw[cg/flow, cgSslow] (m\t) -- (f\n);
    }
    \node[font=\tiny, text=cgSslow, anchor=south] at (0.87, 0.83) {$\bm{1}{-}\va$};
    \draw[cg/flow, cgSslow] (m6) -- (6.75, 0.3);
    \node[cg/tag, anchor=west] at (6.78, 0.3) {next segment};
    \node[cg/dot=cgSslow] (c) at (6.6, 0.3) {};
    \draw[cg/flow, cgSslow, rounded corners=3pt] (c) |- (6.75, 1.05);
    \node[cg/out=cgSslow, anchor=west] (hd) at (6.78, 1.05) {heads\\[-1pt]$\hat{\bm{y}},\, q_{\text{halt}}$};
    \node[font=\tiny, text=cgSslow, anchor=south, align=center] at ($(hd.north) + (0, 0.03)$) {slow channels\\[-2pt]of $\vs$ only};
  \end{scope}
\end{tikzpicture}
\caption{Computation graph of one segment at equal compute (shown for $H = 2$, $n = 2$, i.e.\ $K = 6$, $G = 3$).
(a)~TRM has two learned states, two update rules and two nested loops: the inner loop refines the latent state $\vz$ for $n$ times from the answer $\vy$ and the input $\vx$, and each cycle of the outer loop then updates $\vy$ from $\vz$. Only the last cycle is backpropagated.
(b)~\ours has a single state $\vs$, drawn as one line, and one update rule applied $K$ times: each call mixes the network output $\net(\rms(\vs) + \vx)$ into the state with an exponential moving average, $\vs \gets (\bm{1}-\va)\odot\vs + \va\odot\net(\rms(\vs) + \vx)$, where the share $\bm{1}-\va$ of the previous state bypasses the network (bent arrows) and the rates $a_d$ differ per channel (\cref{fig:mechanisms}a).
The heads read only the slow channels of $\vs$.
Both models have the same number of learnable parameters, the same number of network calls per segment and the same number of differentiated calls.}
\label{fig:graphs}
\end{figure*}
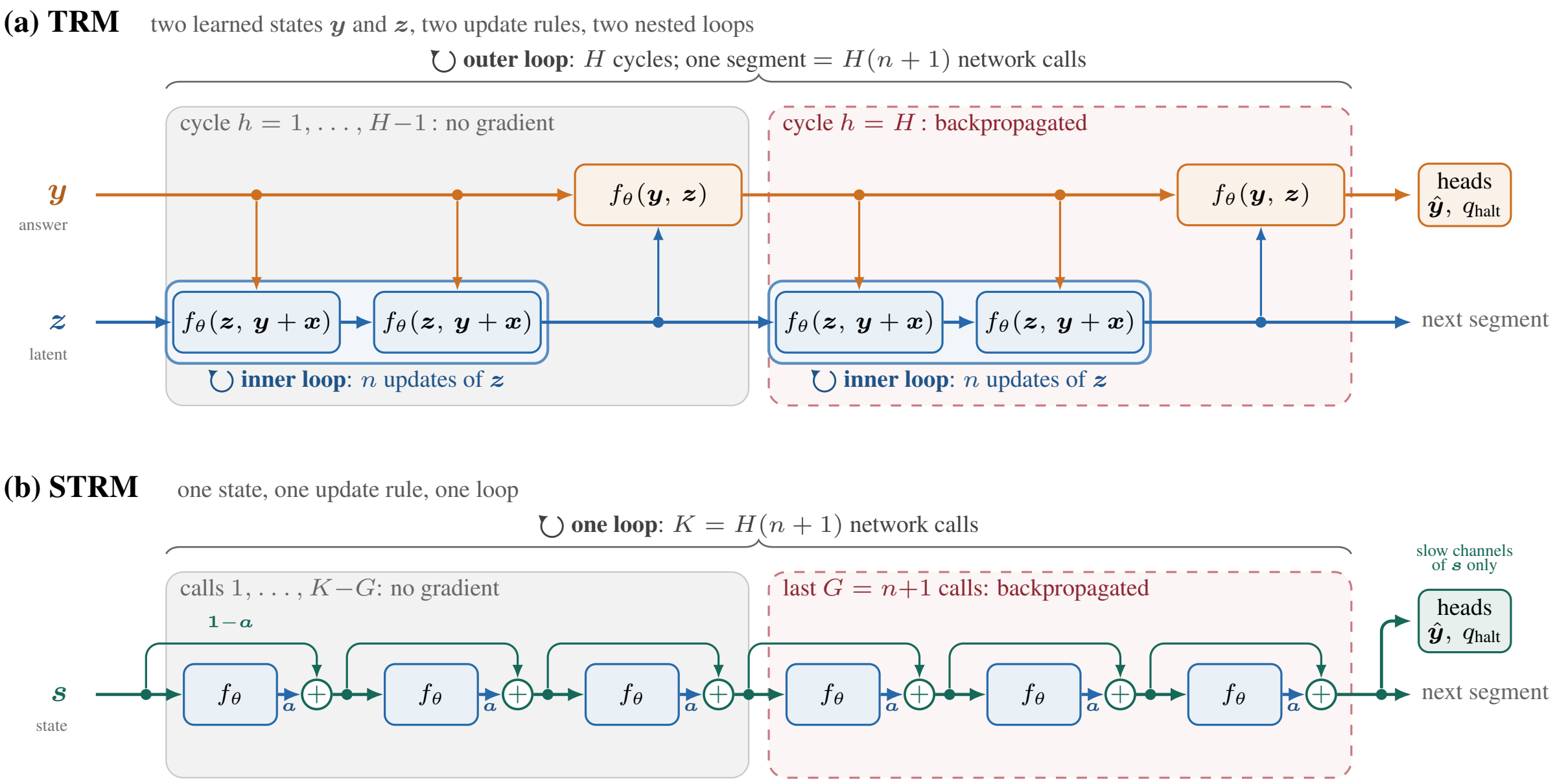

\paragraph{Successors of TRM.}
Most follow-up work keeps TRM's state structure and changes inference or training.
\emph{Test-time scaling.} \Citet{dillon2026speed} use halting time as a confidence signal, reaching $97\%$ on Sudoku-Extreme by selecting the first-halting ensemble member and $96.9\%$ at single-model cost by training four parallel latent states.
The Probabilistic TRM~\citep{sghaier2026ptrm} selects among noisy parallel trajectories with the halting head ($98.75\%$ on Sudoku-Extreme, $91.2\%$ on Pencil Puzzle Bench), and \citet{corbett2026guided} frame such perturbations as approximate inference over latent trajectories.
Inference-time noise escapes initialization-dependent attractors in TRM~\citep{efstathiou2026attractor} and fixed-point trapping in HRM~\citep{ren2026guessing}, and related samplers exist for latent-reasoning language models~\citep{you2025paralleltts}.
\emph{Generative and denoising formulations.} GRAM~\citep{baek2026gram} makes recursive reasoning a latent-variable generative model trained with amortized variational inference.
\Citet{drozdova2026diffusion} iterate a timestep-free diffusion denoiser with a persistent hidden state to arbitrary depth ($99.90\%$ on Sudoku-Extreme).
Looped Flows~\citep{suleymanzade2026loopedflows} train looped models with local denoising objectives, and Flow Reasoning Models~\citep{helbling2026frm} self-condition a flow model on its own outputs and inference-time states.
\emph{Attractors and stability.} Equilibrium Reasoners~\citep{huang2026eqr} train task-conditioned attractors whose fixed points are solutions, the Contraction Mapping Model~\citep{eskin2026dynamical} enforces convergence to a stable equilibrium, and the Recursive Stem Model~\citep{hakimi2026rsm} grows recursion depth with a detached state history and stochastic depth.
\Citet{krishna2026finish} show that running TRM far beyond its training recursion budget keeps raising the exact-solve rate on hard Sudokus.
\emph{Other variants} use Mamba-2 hybrids~\citep{wang2026trmmamba}, reweighted recursion steps~\citep{komisarczyk2026rim}, a depth curriculum~\citep{qasim2025cgar}, or a policy-improvement view~\citep{asadulaev2025policy}.
Their highest accuracies use many parallel trajectories, whereas all our numbers are single-trajectory.

\paragraph{Controlled and damped state updates.}
Closest to our damping mechanism, \citet{shakhvalieva2026limits} compare HRM, TRM and URM in one pipeline, find the hierarchy unnecessary and identify recurrent-state stability as the main bottleneck, and gate the latent update per token, $\vz \leftarrow \operatorname{Norm}(\vz + \bm{\alpha}\odot\Delta)$, a learned single-rate leaky integrator, while keeping both $\vz$ and $\vy$.
In looped language models, Parcae~\citep{prairie2026parcae} stabilizes training with a decaying, negative-diagonal carry, which \citet{buehlmaier2026layernorm} attribute to stabilization alone.
\Citet{sapunov2026utmemory} show that a single-block Universal Transformer fails on Sudoku-Extreme without learned memory tokens, for which our slow averages are a parameter-free alternative.
To our knowledge, no prior HRM or TRM variant uses a bank of multi-timescale averages of the latent \emph{state}.

\paragraph{Multi-timescale recurrence.}
Recurrent networks with several timescales are a classical idea: Clockwork RNNs~\citep{koutnik2014clockwork} update groups of units at different clock periods, hierarchical multiscale RNNs~\citep{chung2017hmrnn} learn when to update each level, and leaky-integrator echo state networks~\citep{jaeger2007leaky} use the update $\bm{x}\leftarrow(1-a)\bm{x} + a\,f(\cdot)$, of which our averages are a fixed-rate, log-spaced instance.
We bring this idea into recursive reasoning models as a replacement for the two-level state.

\section{Method}
\label{sec:method}

We first recall the recursion of the Tiny Recursive Model (TRM)~\citep{jolicoeur2025trm} (\cref{sec:method:trm}) and then introduce \ours, which replaces TRM's two learned states by a single state: a bank of parameter-free exponential moving averages of the network's outputs at many fixed timescales (\cref{sec:method:strm}).
Training follows TRM and is summarized in \cref{sec:method:training}.

\paragraph{Notation.}
A puzzle is a token sequence of length $L$ (a flattened grid), embedded together with $P$ learned register tokens into $\vx \in \mathbb{R}^{(P+L)\times D}$.
All states have the same shape as $\vx$.
$\net$ denotes the single shared network: given a state $\vz$ and an injection $\bm{u}$ it returns $\net(\vz,\bm{u}) = B_2(B_1(\vz + \bm{u}))$, and we write $\net(\bm{u}) = \net(\bm{0},\bm{u}) = B_2(B_1(\bm{u}))$ when there is no state input, with two post-norm residual blocks $B_1,B_2$ (token mixing followed by a SwiGLU MLP~\citep{shazeer2020glu}, each with RMSNorm~\citep{zhang2019rmsnorm}).
Token mixing is a self-attention layer with rotary position embeddings~\citep{su2024roformer} or, for Sudoku, an MLP over the sequence axis, as in TRM.
$\rms(\cdot)$ normalizes over the channel axis, and $\odot$ is the element-wise product.

\subsection{Background: the Tiny Recursive Model}
\label{sec:method:trm}

TRM keeps two states, both updated by the same network:
\begin{description}
  \item[$\vy$:] the \emph{answer} state from which output is decoded (HRM's slow high-level state meant for abstract planning),
  \item[$\vz$:] the \emph{latent} state, the reasoning that improves $\vy$ (HRM's fast low-level state meant for detailed computation).
\end{description}
In each cycle, the latent state is refined $n$ times from the input and the current answer, and the answer is then updated from the refined latent state.
One \emph{segment} repeats this cycle $H$ times (\cref{alg:trm}) and therefore applies the network $H(n+1)$ times in total.

\begin{algorithm2e}[H]
\caption{One segment of the TRM recursion.}
\label{alg:trm}
\KwIn{input embedding $\vx$, carried states $(\vy,\vz)$, cycles $H$, latent updates per cycle $n$}
\For{$h = 1,\dots,H$}{
  \tcp{without gradient for $h < H$}
  \For{$i = 1,\dots,n$}{
    $\vz \gets \net(\vz,\, \vy + \vx)$\;
  }
  $\vy \gets \net(\vy,\, \vz)$\;
}
\KwRet{$\vy,\vz$}
\end{algorithm2e}

Only the last cycle is differentiated. 
The first $H-1$ cycles run without gradient and both states are detached between segments.
The output head decodes $\vy$ token-wise and a halting head reads the first register position of $\vy$ to determine whether to stop the recursion.
The deep supervision and adaptive halting of TRM is described together with the training of \ours in \cref{sec:method:training}.

The two-state approach is necessary in TRM, since ablations in~\citep{jolicoeur2025trm} show that a plain direct update with a single state overfits (in our runs more than $99\%$ training accuracy but only $52.5$--$57.4\%$ test accuracy at the end of training, \cref{tab:sudoku-ablation}).
We argue that the role of $\vy$ in TRM is regularization through a slow and delayed update.
Reading off the solution happens only through $\vy$ as well, acting as another information bottleneck for regularization purposes.

\subsection{\ours}
\label{sec:method:strm}

\ours keeps TRM's network, deep supervision and halting,
while it replaces its two-level recursion by a single state $\vs$ with one update rule in one loop (\cref{fig:graphs}).
The state is updated through exponential moving averages with fixed, channel-dependent rates.
Implicitly, fast channels act similar to the latent state $\vz$ and slow channels to the answer state $\vy$.
With $K = H(n+1)$ calls per segment, \ours has the same parameters and compute as TRM.

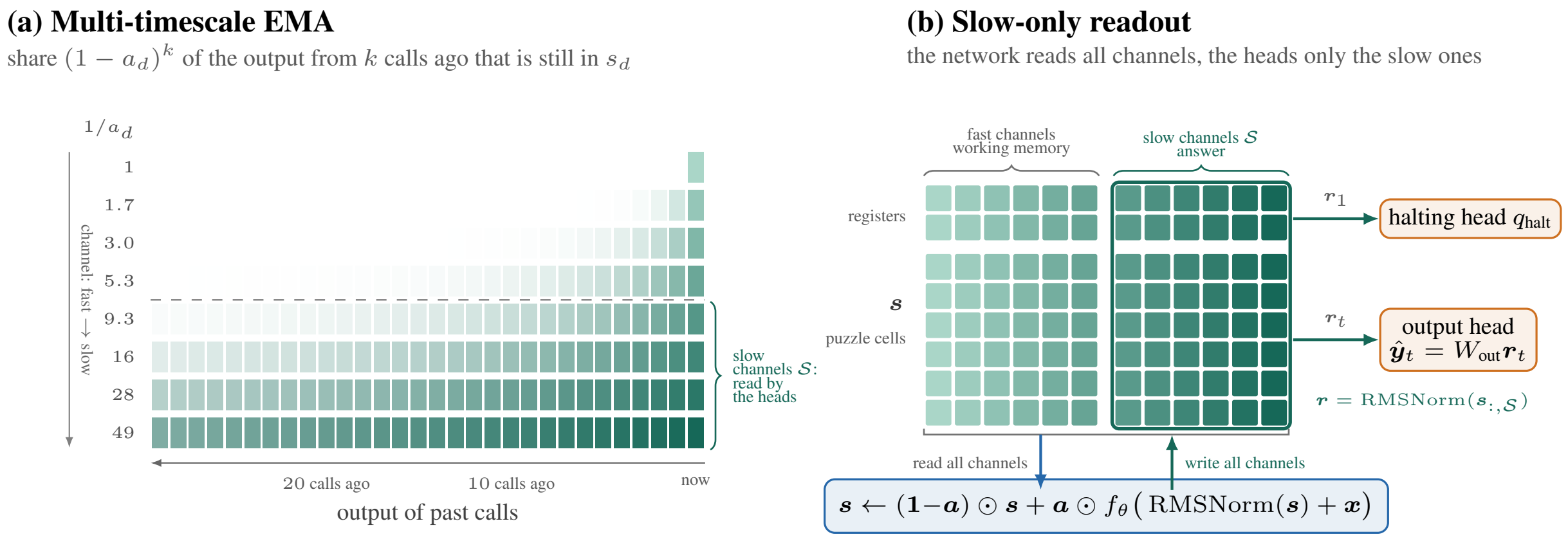
\begin{figure*}[t]
\centering
\definecolor{mcZ}{RGB}{38,104,170}
\definecolor{mcSfast}{RGB}{170,214,200}
\definecolor{mcSslow}{RGB}{22,104,88}
\definecolor{mcY}{RGB}{206,112,32}
\tikzset{
  mc/title/.style={anchor=west, font=\small\bfseries},
  mc/sub/.style={anchor=west, font=\scriptsize, text=black!65},
  mc/tick/.style={font=\tiny, text=black!65},
  mc/label/.style={font=\scriptsize, text=black!80},
  mc/arrow/.style={-{Latex[length=1.6mm, width=1.3mm]}, line width=0.8pt},
  mc/out/.style={draw=#1, fill=#1!10, rounded corners=3pt, font=\scriptsize, align=center,
    inner sep=2.5pt, line width=0.6pt},
  mc/net/.style={draw=mcZ, fill=mcZ!10, rounded corners=3pt, line width=0.6pt, font=\scriptsize,
    minimum height=0.55cm, inner xsep=4pt},
}
\begin{tikzpicture}
  \node[mc/title] at (-0.3, 4.15) {(a) Multi-timescale EMA};
  \node[mc/sub] at (-0.3, 3.8) {share $(1 - a_d)^k$ of the output from $k$ calls ago that is still in $s_d$};
  \begin{scope}[xshift=1.3cm, yshift=-0.25cm, x=0.19cm, y=0.39cm]
  \colorlet{erSfast}{mcSfast}\colorlet{erSslow}{mcSslow}
    \pgfmathsetmacro{\amn}{1/49}
    \foreach \r/\tau in {0/1, 1/1.7, 2/3.0, 3/5.3, 4/9.3, 5/16, 6/28, 7/49} {
      \pgfmathsetmacro{\a}{\amn^(\r/7)}
      \pgfmathsetmacro{\p}{\r * 100 / 7}
      \colorlet{rowc}{erSslow!\p!erSfast}
      \pgfmathsetmacro{\y}{7 - \r}
      \foreach \k in {0, ..., 29} {
        \pgfmathsetmacro{\w}{100 * (1 - \a)^\k}
        \fill[rowc!\w!white] (29 - \k + 0.08, \y + 0.1) rectangle (29 - \k + 0.92, \y + 0.9);
      }
      \node[font=\tiny, text=black!70, anchor=east] at (-0.2, \y + 0.5) {$\tau$};
    }
    \node[font=\tiny, text=black!70, anchor=south east] at (-0.2, 7.95) {$1/a_d$};
    \draw[-{Latex[length=1.1mm]}, black!60, line width=0.4pt] (30, -0.3) -- (0, -0.3);
    \node[font=\tiny, text=black!65, anchor=north] at (29.5, -0.35) {now};
    \foreach \k in {10, 20} { \node[font=\tiny, text=black!65, anchor=north] at (29.5 - \k, -0.35) {$\k$ calls ago}; }
    \node[font=\scriptsize, text=black!75, anchor=north] at (15, -1.05) {output of past calls};
    \draw[-{Latex[length=1.1mm]}, black!50, line width=0.4pt] (-4.4, 7.9) -- node[sloped, above, font=\tiny, text=black!60] {channel: fast $\rightarrow$ slow} (-4.4, 0.1);
    \draw[decorate, decoration={brace, amplitude=3pt, raise=1pt}, erSslow, line width=0.5pt] (30.1, 3.95) -- node[right=4pt, font=\tiny, text=erSslow, align=left] {slow\\[-2pt]channels $\mathcal{S}$:\\[-2pt]read by\\[-2pt]the heads} (30.1, 0.05);
    \draw[black!55, dashed, line width=0.4pt] (0, 4) -- (30, 4);
  \end{scope}

  \begin{scope}[xshift=9.25cm, x=0.3cm, y=0.3cm]
    \node[mc/title] at (-1, 13.83) {(b) Slow-only readout};
    \node[mc/sub] at (-1, 12.67) {the network reads all channels, the heads only the slow ones};
    \foreach \c in {0, ..., 11} {
      \pgfmathsetmacro{\p}{\c * 100 / 11}
      \pgfmathsetmacro{\xx}{\c + (\c > 5 ? 0.5 : 0)}
      \foreach \r in {0, ..., 7} {
        \pgfmathsetmacro{\yy}{\r + (\r > 5 ? 0.35 : 0)}
        \fill[mcSslow!\p!mcSfast, rounded corners=0.6pt] (\xx + 0.06, \yy + 0.06) rectangle (\xx + 0.94, \yy + 0.94);
      }
    }
    \node[mc/label, anchor=east] at (-0.3, 4.2) {$\vs$};
    \node[mc/tick, anchor=east] at (-0.2, 7.2) {registers};
    \node[mc/tick, anchor=east] at (-0.2, 3) {puzzle cells};
    \draw[decorate, decoration={brace, amplitude=3pt, raise=1pt}, black!55, line width=0.5pt] (0, 8.5) -- node[above=3pt, mc/tick, align=center] {fast channels\\[-2pt]working memory} (6, 8.5);
    \draw[decorate, decoration={brace, amplitude=3pt, raise=1pt}, mcSslow, line width=0.5pt] (6.5, 8.5) -- node[above=3pt, mc/tick, text=mcSslow, align=center] {slow channels $\mathcal{S}$\\[-2pt]answer} (12.5, 8.5);
    \draw[mcSslow, line width=1pt, rounded corners=2pt] (6.45, -0.05) rectangle (12.55, 8.4);
    \node[mc/net] (f) at (6.25, -2.7) {$\vs \gets (\bm{1}{-}\va)\odot\vs + \va\odot\net\big(\rms(\vs) + \vx\big)$};
    \draw[black!55, line width=0.5pt] (0, -0.05) -- (0, -0.25) -- (12.5, -0.25) -- (12.5, -0.05);
    \draw[mc/arrow, mcZ] (4, -0.25) -- node[left, mc/tick] {read all channels} (4, -2.15);
    \draw[mc/arrow, mcSslow] (8.5, -2.15) -- node[right, mc/tick, text=mcSslow] {write all channels} (8.5, -0.25);
    \node[mc/out=mcY, anchor=west] (halt) at (15.6, 7.15) {halting head $q_{\text{halt}}$};
    \node[mc/out=mcY, anchor=west, align=center] (outh) at (15.6, 3) {output head\\[-1pt]$\hat{\bm{y}}_t = W_{\text{out}}\vr_t$};
    \draw[mc/arrow, mcSslow] (12.65, 7.15) -- node[above, mc/tick] {$\vr_1$} (halt.west);
    \draw[mc/arrow, mcSslow] (12.65, 3) -- node[above, mc/tick] {$\vr_t$} (outh.west);
    \node[mc/tick, anchor=north west, align=left, text=mcSslow] at (13.0, 1.6) {$\vr = \rms(\vs_{:,\mathcal{S}})$};
  \end{scope}
\end{tikzpicture}
\caption{The state of \ours.
(a)~Multi-timescale EMA: unrolling the update gives $s_d = \sum_{k \ge 0} a_d (1 - a_d)^k\, o_d^{(-k)}$, where $o_d^{(-k)}$ is the network output of $k$ calls ago. The shade shows its relative weight $(1 - a_d)^k$ for $8$ of the $D = 512$ channels, with rates log-spaced from $1$ to $\amin = 1/49$. The fastest channel holds only the latest output, while the slowest still holds a large share of outputs from $30$ calls ago.
(b)~Slow-only readout: in every call the network reads and writes all channels of $\vs$, but the output head (puzzle tokens) and the halting head (first register token) read only the RMS-normalized slow channels $\mathcal{S} = \{d : a_d \le R\}$, so the fast channels act as a working memory that reaches the loss only through later calls.}
\label{fig:mechanisms}
\end{figure*}

\paragraph{Multi-timescale state.}
\label{sec:method:flat}
\ours carries a single state $\vs$ whose channels are parameter-free exponential moving averages (EMAs) of the network's outputs, with the fixed rates
\begin{equation}
  a_d \;=\; \amax \left(\frac{\amin}{\amax}\right)^{\frac{d-1}{D-1}},
  \label{eq:rates}
\end{equation}
log-spaced from $\amax = 1$ to $\amin$ (a time constant of $1/\amin$ network calls) and sorted from fast to slow along the channel axis (\cref{fig:mechanisms}a).
\Cref{alg:flat} shows one segment of $K$ calls, of which the first $K-G$ run without gradient.

\begin{algorithm2e}[H]
\caption{One segment of the \ours recursion.}
\label{alg:flat}
\KwIn{input embedding $\vx$, carried state $\vs$, rates $\va$, calls $K$, differentiated calls $G$}
\For{$k = 1,\dots,K$}{
  \tcp{without gradient for $k \le K-G$}
  $\vs \gets (\bm{1}-\va) \odot \vs + \va \odot \net\big(\rms(\vs) + \vx\big)$\;
}
\KwRet{$\vs$}
\end{algorithm2e}

Each network call computes and mixes the output $\vo = \net(\rms(\vs) + \vx)$ into the state at the per-channel rates.
Channels with rates close to $1$ copy most of the latest output and act as working memory, while the slow channels with rates close to $0$ aggregate outputs over many calls.
We use RMSNorm without gain to bound the input of the network and stabilize the updates of $\vs$.
To match TRM's compute, we set $K = H(n+1)$ and $G = n+1$.
The state is carried across segments (detached) and reset to a fixed random vector $\vs_0$, shared by all tokens, when a new puzzle enters the slot.

\paragraph{Slow-only readout.}
\label{sec:method:readout}
If the heads read all channels of $\vs$, including the rate-$1$ channels that copy the latest output, the loss can be reduced through the fast path alone.
Instead, the output and halting heads read \emph{only the slow channels} (\cref{fig:mechanisms}b) $\mathcal{S} = \{d : a_d \le R\}$ of $\vs$, for a cut-off rate $R$:
\begin{equation}
  \begin{aligned}
    \vr &= \rms(\vs_{:,\mathcal{S}}), \\
    \hat{\bm{y}}_t &= W_{\text{out}}\, \vr_t, \quad t = P+1,\dots,P+L, \\
    q_{\text{halt}} &= \bm{w}_{q}^{\top} \vr_1 + b_q .
  \end{aligned}
  \label{eq:readout}
\end{equation}
Here $\vr_t$ is the row of token $t$, where tokens $1,\dots,P$ are the registers and $P+1,\dots,P+L$ the puzzle cells.
As in TRM (\cref{sec:method:trm}), the output head $W_{\text{out}}$ decodes every puzzle token and the halting head reads the first register token, now of $\vr$ instead of $\vy$.
This reproduces TRM's split between $\vz$ and $\vy$ inside the single state $\vs$: whatever the network computes in its fast channels must be written into channels that average over many calls before it can affect the output.
We use $R = \sqrt{\amin}$, which selects the slower half of the channels ($256$ of $D=512$).
\Cref{sec:experiments:ablation} compares against heads that read all channels.

\paragraph{Rate jitter.}
\label{sec:method:jitter}
As an additional regularization, we randomly perturb the rates during training.
The size of the perturbation is determined by the \emph{jitter strength} $J \ge 1$.
Whenever a new puzzle enters a batch slot, we draw an exponent $c$ and replace every rate $a_d$ by a perturbed rate $\tilde{a}_d$,
\begin{equation}
  \begin{aligned}
    u &\sim \mathcal{U}[-1, 1], \qquad c = J^{\,u} \in [1/J,\, J], \\
    \tilde{a}_d &= a_d^{\,c} = \big(\amin^{\,c}\big)^{\frac{d-1}{D-1}},
  \end{aligned}
  \label{eq:jitter}
\end{equation}
and run all segments of this puzzle with the rates $\tilde{\va}$ instead of $\va$ in \cref{alg:flat}.
The perturbed rates are still log-spaced, now from $1$ to $\amin^{\,c}$, so the channels keep their order and the rate-$1$ channels still copy the latest output $\vo$.
In terms of time constants, $1/\tilde{a}_d = (1/a_d)^{c}$, so all channels become slower for $c > 1$ and faster for $c < 1$.
We use $J = 1.5$, so that for $\amin = 1/49$ the slowest time constant $1/\amin^{\,c}$ varies between $49^{2/3} \approx 13$ and $49^{3/2} = 343$ calls.
The jitter is meant as a regularizer: the model can rely on which channels are fast and which are slow, but not on their exact time constants.
At evaluation we use the nominal rates ($c = 1$).

\subsection{Training}
\label{sec:method:training}

Apart from the recursion, TRM and \ours are trained identically, following TRM.
\begin{description}
  \item[Deep supervision.]
  The recursion on a puzzle continues for up to $N_{\text{sup}} = 16$ segments, and the states are carried over from segment to segment (detached).
  Every segment receives the full loss and its own optimizer step.
  \item[Loss.]
  A stable-max cross-entropy on the puzzle tokens trains the decoded solution.
  A binary cross-entropy, weighted by $0.5$, trains the halting logit $q_{\text{halt}}$ (\cref{eq:readout}) to predict whether the current answer is exactly correct.
  \item[Adaptive halting.]
  In the style of ACT~\citep{graves2016act}, training on a puzzle stops once $q_{\text{halt}} > 0$.
  With exploration probability $p = 0.1$, it stops only after running a random minimum number of segments.
  \item[Batch slots.]
  A batch consists of a fixed number of slots, each holding one puzzle, and every optimizer step runs one segment on every slot.
  When a puzzle stops, it is replaced in its slot by a new puzzle with reset states (and, for \ours, a new jitter exponent $c$).
  A batch therefore mixes puzzles at different stages of their recursion.
  \item[Evaluation.]
  All $N_{\text{sup}}$ segments are run without halting, with an exponential moving average of the model weights~\citep{polyak1992averaging} (decay $0.999$).
\end{description}

\section{Experiments}
\label{sec:experiments}

We compare \ours with TRM on four benchmarks, on ARC-AGI against published results and otherwise under identical data, optimizer (up to the learning rate and weight decay stated below), deep supervision and evaluation, at equal compute per segment and with the same number of parameters.
We also include other methods that try to improve upon TRM.
If not stated otherwise, all numbers are exact accuracy (a puzzle counts only if every cell is correct), measured with the EMA weights after $N_{\text{sup}} = 16$ segments.

\begin{table*}[!t]
\centering
\caption{Sudoku-Extreme, exact accuracy (\%) on the full test set ($422{,}786$ puzzles), two seeds.
\emph{Best} denotes the checkpoint with the highest accuracy on a subset of test puzzles and \emph{final} the weights after the last step.
FPRM~\citep{movahedi2026fprm}, SFPRM~\citep{zela2026sfprm} and RSM~\citep{hakimi2026rsm} are trained with the authors' code and hyperparameters on our data and budget (\cref{sec:experiments:sudoku}). \\ $^\dagger$Peak learning rate of the authors' recipe.}
\label{tab:sudoku}
\begin{tabular}{llcccc}
\toprule
 & & \multicolumn{2}{c}{best} & \multicolumn{2}{c}{final} \\
\cmidrule(lr){3-4}\cmidrule(lr){5-6}
model & peak lr & seeds 0 / 1 & mean & seeds 0 / 1 & mean \\
\midrule
TRM & $10^{-4}$ & 88.4 / 88.1 & 88.3 & 70.0 / 84.6 & 77.3 \\
TRM & $1.5\!\times\!10^{-4}$ & 87.9 / 87.1 & 87.5 & 82.6 / 69.8 & 76.2 \\
FPRM & $10^{-3}$\,$^\dagger$ & 66.1 / 76.1 & 71.1 & 63.5 / 75.5 & 69.5 \\
SFPRM & $5\!\times\!10^{-3}$\,$^\dagger$ & 74.1 / 73.0 & 73.6 & 12.2 / 17.1 & 14.7 \\
RSM & $10^{-3}$\,$^\dagger$ & 20.3 / 20.6 & 20.5 & 20.3 / 18.3 & 19.3 \\
\midrule
\ours & $1.5\!\times\!10^{-4}$ & \textbf{91.3} / \textbf{93.0} & \textbf{92.1} & \textbf{90.3} / \textbf{92.1} & \textbf{91.2} \\
\bottomrule
\end{tabular}
\end{table*}

\begin{figure*}[t]
\centering
\includegraphics[width=\textwidth]{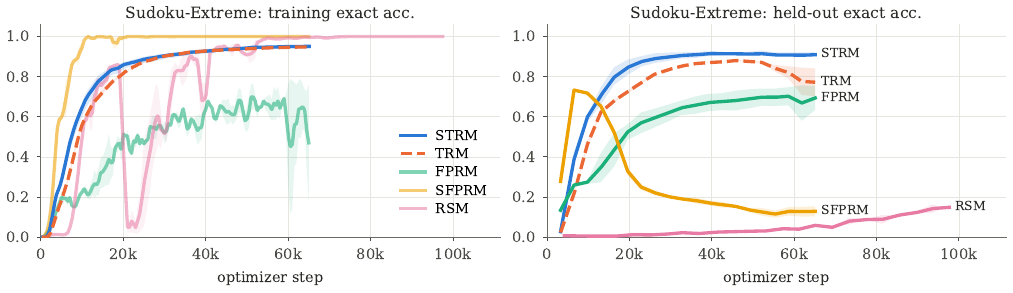}
\caption{Convergence on Sudoku-Extreme: training and held-out exact accuracy over the optimizer steps, mean over the seeds of \cref{tab:sudoku} with the range shaded.
TRM uses peak learning rate $10^{-4}$.
SFPRM and RSM solve every training puzzle while their held-out accuracy stays below $73\%$, TRM's held-out accuracy declines late in training, while that of \ours holds.
}
\label{fig:convergence-baselines}
\end{figure*}

\begin{figure*}[!t]
\centering
\includegraphics[width=\textwidth]{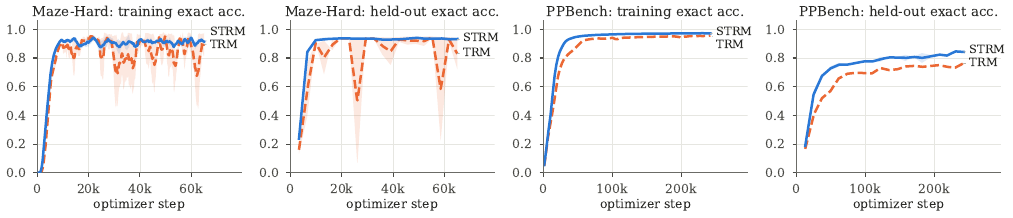}
\caption{Convergence on Maze-Hard and PPBench: training and held-out exact accuracy over the optimizer steps, mean over two seeds, except for TRM on PPBench (one seed).
The held-out accuracy is measured on the test sets ($1{,}000$ mazes and $393$ PPBench puzzles).}
\label{fig:convergence-maze-ppbench}
\end{figure*}

\paragraph{Training recipe and hyperparameters.}
To keep hyperparameter tuning from deciding the comparison, both models, including our TRM baseline, use a modified version of the original TRM training recipe~\citep{jolicoeur2025trm}.
We replace the Adam-atan2 optimizer of the original code by AdamW, apply weight decay only to weight matrices instead of all parameters, and clip the gradient norm at $1.0$, which the original code does not do.
On Sudoku-Extreme we also decay the learning rate with a cosine schedule, whereas the original recipe keeps it constant after warm-up.
With these changes, TRM reaches $88.3\%$ at the best checkpoint on Sudoku-Extreme (\cref{tab:sudoku}), compared to the $87.4\%$ reported by \citet{jolicoeur2025trm}.
We also swept the hyperparameters of TRM thoroughly, so our TRM baseline is at least as well tuned as \ours.
Apart from the recursion, the models differ only in the peak learning rate ($10^{-4}$ for TRM, $1.5\times10^{-4}$ for \ours), the weight decay, tuned per model, and on Sudoku-Extreme the final learning rate (\cref{sec:experiments:sudoku}).
On Sudoku-Extreme we also trained TRM at $1.5\times10^{-4}$, and it did not improve (\cref{tab:sudoku}).

\subsection{Sudoku-Extreme}
\label{sec:experiments:sudoku}

We use the Sudoku-Extreme data of \citet{wang2025hrm}: $1{,}000$ training puzzles, each with $1{,}000$ augmented copies, and $422{,}786$ test puzzles.
Both models use the MLP token mixer of TRM (TRM-MLP), with hidden size $D=512$, two blocks and $P=16$ register tokens.
TRM uses $H=3$ and $n=6$.
\ours uses $K = 21$ calls per segment with gradient through the last $G = 7$, rates from $1$ to $\amin = 1/49$, readout cut-off $R = 1/7$ and jitter $J=1.5$.
Training runs for $50{,}000$ epochs ($65{,}104$ optimizer steps, batch $768$) with AdamW~\citep{loshchilov2019adamw} ($\beta = (0.9, 0.95)$, weight decay $1.0$ on matrices, gradient clipping at $1.0$), $2{,}000$ warm-up steps and cosine decay to zero.
\ours uses weight decay $0.7$ and decays the learning rate to $0.1\times$ its peak instead of zero, and with $D = 512$ it has $5.03$M parameters, same as TRM.
We train FPRM~\citep{movahedi2026fprm}, SFPRM~\citep{zela2026sfprm} and RSM~\citep{hakimi2026rsm} with the authors' code and hyperparameters on the same training data and budget.
SFPRM uses Muon for the weight matrices and AdamW with peak learning rate $10^{-3}$ for the remaining parameters; FPRM and RSM use AdamW.
During training we evaluate every $3{,}255$ steps on the first $10{,}368$ test puzzles.
We report the full test set results both for the checkpoint with the best value on this subset (\emph{best}) and for the final weights (\emph{final}).
Since the subset is part of the test set, \emph{best} carries a small selection bias, whereas \emph{final} does not.

\begin{table*}[!t]
\centering
\caption{PPBench, exact accuracy (\%) on the test set ($393$ puzzles) and on the golden set ($61$ puzzles).
\emph{Late}: mean over the twelve evaluations at steps $109$k--$242$k, also per puzzle type, and \emph{best} and \emph{final}: the best and the last evaluation (\cref{sec:experiments:ppbench}).
\emph{Average}: unweighted mean of the six types, whereas the test columns weigh each type by its number of puzzles.
\ours: mean over two seeds, TRM: one seed.}
\label{tab:ppbench}
\small
\setlength{\tabcolsep}{3.4pt}
\begin{tabular}{lccc|ccccccc|ccc}
\toprule
 & \multicolumn{3}{c|}{test} & \multicolumn{7}{c|}{test, late, per type} & \multicolumn{3}{c}{golden} \\
model & late & best & final & heyawake & lightup & nurikabe & shakashaka & sudoku & tapa & average & late & best & final \\
\midrule
TRM (our implementation) & 73.4 & 76.3 & 76.3 & 77.7 & 93.7 & 80.6 & 96.1 & 41.4 & 55.3 & 74.1 & 60.7 & 67.2 & 57.4 \\
\ours (2 seeds) & \textbf{81.0} & \textbf{84.7} & \textbf{84.2} & \textbf{90.8} & \textbf{97.9} & \textbf{88.7} & \textbf{96.4} & \textbf{50.3} & \textbf{69.0} & \textbf{82.2} & \textbf{67.6} & \textbf{74.6} & \textbf{72.2} \\
\bottomrule
\end{tabular}
\end{table*}

\begin{table}[!t]
\centering
\caption{Maze-Hard, exact accuracy (\%) on the $1{,}000$ test mazes, seeds 0 / 1.
\emph{Late}: mean over the eleven evaluations at steps $32{,}550$--$65{,}100$.}
\label{tab:maze}
\setlength{\tabcolsep}{3pt}
\begin{tabular}{lccc}
\toprule
model & best & final & late  \\
\midrule
TRM & 94.0 / \textbf{95.0} & 72.1 / \textbf{93.8} & 80.7 / \textbf{94.6}  \\
\ours & \textbf{95.1} / 93.3 & \textbf{94.3} / 92.2 & \textbf{94.0} / 92.7 \\
\bottomrule
\end{tabular}
\end{table}


\Cref{tab:sudoku} shows the main result.
\ours is the most accurate model: its final weights are $2.9$ points above the best checkpoint of TRM and $13.9$ points above the final weights of TRM, since TRM's held-out accuracy collapses late in training (\cref{sec:experiments:stability}) while \ours stays within one point of its best checkpoint.
The fixed-point and recursive models fall far behind: the best of them at its peak, SFPRM, is $17.6$ points below the final weights of \ours and then declines, and the best of them with the final weights, FPRM, trails by $21.7$ points.
SFPRM and RSM fit the training set almost perfectly while generalizing poorly (\cref{fig:convergence-baselines}).
None of them seems to work on the restricted training set of $1{,}000$ puzzles used by HRM and TRM. 
Their published results rely on more data or more compute.
SFPRM reports $96.6\%$ after training on $5{,}000$ distinct puzzles, five times as many as here, and also collapses late in training on the $1{,}000$-puzzle build~\citep{zela2026sfprm}.
RSM reports $97.5\%$ with additional test-time recursion after training on the full training split of $3.8$M distinct puzzles without augmentation~\citep{hakimi2026rsm}.
FPRM reports $94.2\%$ on the same $1{,}000$-puzzle build, but with $60{,}000$ epochs and up to $35{,}000$ network applications per test puzzle, about $100$ times our budget~\citep{movahedi2026fprm}. \citet{zela2026sfprm} find that FPRM retrained on this build stays well below that number.

\subsection{PPBench}
\label{sec:experiments:ppbench}

PPBench~\citep{waugh2026ppbench} contains pencil puzzles of many types.
We use the six types heyawake, lightup, nurikabe, shakashaka, sudoku and tapa at grid size $10\times 10$ ($L = 100$, a joint vocabulary of $675$ tokens).
The training set has $9{,}305$ puzzles, each with $10$ augmented copies (partially solved states under random dihedral transformations).
We evaluate on a test set of $393$ puzzles (heyawake $72$, lightup $54$, nurikabe $67$, shakashaka $75$, sudoku $95$, tapa $30$) and on the $61$ puzzles of PPBench's \emph{golden} set at these types and size.
Both models use self-attention with 1D rotary embeddings (TRM-Att), $D=512$, two blocks, $8$ heads and $P=16$ registers.
TRM uses $H=3$ and $n=4$.
\ours uses $K = 15$, $G = 5$ and $J=1.5$.
Training runs for $20{,}000$ epochs ($242{,}300$ steps, batch $768$) with the same AdamW settings as for Sudoku-Extreme, $2{,}000$ warm-up steps and a \emph{constant} learning rate thereafter (cosine decay to zero overfits strongly on this benchmark in the baseline experiments).
We evaluate every $12{,}115$ steps.
Because the test and golden sets are small (the standard error of a single evaluation is about $2.2$ and $6.3$ points), we report the mean over the twelve evaluations in the second half of training (steps $109$k--$242$k, \emph{late mean}) in addition to the best and final evaluations.

\Cref{tab:ppbench} shows the results.
\ours is the best model on the test set, $7.6$ points above TRM ($8.4$ points at the best evaluation), and on the golden set ($+6.9$ points), and it leads on every puzzle type.

\begin{table}[!t]
\centering
\caption{ARC-AGI benchmarks, exact accuracy (\%) on the $400$ test samples for ARC-AGI-1, and $120$ for ARC-AGI-2.
Each model is allowed two attempts (pass@2) to solve the geometric puzzles,
and the accuracy is averaged on all tasks.}
\label{tab:arc}
\setlength{\tabcolsep}{3pt}
\begin{tabular}{lcc}
\toprule
model & ARC-AGI-1 & ARC-AGI-2 \\
\midrule
TRM & 44.6 & 7.8 \\
FPRM & 47.5 & 6.2 \\
SFPRM & 47.4 & 3.8 \\
\ours & \textbf{50.0} & \textbf{8.3} \\
\bottomrule
\end{tabular}
\end{table}

\subsection{Maze-Hard}
\label{sec:experiments:maze}

We use the Maze-Hard data of \citet{wang2025hrm}: $1{,}000$ training mazes of size $30\times 30$, each under $8$ dihedral transformations with the shortest paths recomputed, and $1{,}000$ test mazes.
Both models use the attention setup of PPBench (TRM-Att).
TRM uses $H=3$ and $n=4$, and \ours $K = 15$, $G = 5$, $\amin = 1/13$ and $R = 1/\sqrt{13}$.
Training runs for $65{,}100$ steps at batch $768$ with a constant learning rate after warm-up, and we evaluate every $3{,}255$ steps on the full test set.
TRM uses learning rate $10^{-4}$ and weight decay $1.0$, and \ours learning rate $1.5\times10^{-4}$ and weight decay $0.7$.
Both models have $6.82$M parameters.

Note that we have corrected augmentations as compared to TRM's setup.
TRM's code augments a maze by applying the dihedral transformation to the original shortest path.
Shortest paths are in general not unique, and the original paths follow the tie-breaking rule of the search that generated them, which a transformed path violates.
The transformed labels are therefore inconsistent with the test set. 
TRM trained on them drops from $84.6\%$ without augmentation to $11.7\%$.
We recompute the path of every transformed maze with the original tie-breaking rule.
With this correction, our TRM baseline reaches $94.5\%$ at its best checkpoint (\cref{tab:maze}), compared
 to the published $85.3\%$~\citep{jolicoeur2025trm}.

On Maze-Hard, \ours and TRM reach around the same accuracy at their best checkpoints ($94.2\%$ and $94.5\%$, \cref{tab:maze}).
\ours learns much faster and keeps its accuracy until the end of training, whereas TRM is less stable (\cref{fig:convergence-maze-ppbench}).

\subsection{ARC-AGI}
\label{sec:experiments:arc}


The abstract reasoning benchmarks ARC-AGI-1~\citep{chollet2019measure}
and ARC-AGI-2~\citep{chollet2025arcagi2} are designed to test the model's ability to solve new tasks from minimal demonstrations.
ARC-AGI-2 is harder than ARC-AGI-1, with more complex problems.
Following TRM, we include the related ConceptARC dataset~\citep{moskvichev2023conceptarc} as additional training instances and use identical training and test data splits and augmentations.
The architecture is identical to that on PPBench, and training runs for a total of $100{,}000$ epochs, as did TRM.
TRM uses learning rate $10^{-4}$ and weight decay $0.1$, and \ours learning rate $1.5\times10^{-4}$, weight decay $0.1$, $\amin = 1/13$ and $R = 1/\sqrt{13}$.

On ARC-AGI, \ours achieves $50.0\%$ pass@2 accuracy on ARC-AGI-1 and $8.3\%$ on ARC-AGI-2, $5.4$ and $0.5$ points above the published TRM results and $2.5$ and $0.5$ points above the best of TRM, FPRM and SFPRM, which use the same two-block network, see Table~\ref{tab:arc}.
The numbers for other methods are taken from the original papers.
At approximately $50\%$ training epochs on ARC-AGI-1 and $60\%$ on ARC-AGI-2,
\ours surpasses TRM's best performance, and is still improving till the final step.
There are approaches like URM~\citep{gao2025urm} that report higher accuracies ($53.8\%$ and $16.0\%$ pass@1), but typically are not matched regarding training. For example, URM uses four instead of two blocks and twice the epochs.

\subsection{Ablations}
\label{sec:experiments:ablation}

\begin{table}[!t]
\centering
\caption{Sudoku-Extreme ablations of \ours, exact accuracy (\%) of the final weights on the full test set and on the training batches (online weights, last $5\%$ of the steps, mean over the seeds).
Each row changes one setting of the reference configuration ($\amin = 1/49$, $R = 1/7$, $J = 1.5$, weight decay $0.7$, $D = 512$).
The rows on the slowest rate use weight decay $1.0$ and one seed, and the last rows run the reference weights with different test-time segments ($16$ in training and in all other rows, including the reference).
}
\label{tab:sudoku-ablation}
\setlength{\tabcolsep}{4pt}
\begin{tabular}{lccc}
\toprule
 & train & \multicolumn{2}{c}{test} \\
\cmidrule(lr){2-2}\cmidrule(lr){3-4}
variant & mean & seeds 0 / 1 & mean \\
\midrule
reference & 95.6 & 90.3 / 92.1 & \textbf{91.2} \\
no averaging (all rates $1$) & 99.3 & 57.4 / 52.5 & 55.0 \\
single rate $1/7$ & 96.1 & 82.6 / 81.2 & 81.9 \\
heads read all channels & 97.1 & 83.4 / 90.5 & 86.9 \\
no rate jitter ($J = 1$) & 95.6 & 90.8 / 90.5 & 90.7 \\
stronger rate jitter ($J = 2$) & 96.5 & 90.5 / 90.6 & 90.6 \\
\midrule
$\amin = 1/13$ & 94.0 & 87.1 & -- \\
$\amin = 1/75$ & 93.6 & 90.0 & -- \\
$\amin = 1/100$ & 93.3 & 89.9 & -- \\
\midrule
$8$ test-time segments & -- & 84.0 / 86.3 & 85.1 \\
$32$ test-time segments & -- & 94.1 / 95.5 & 94.8 \\
$64$ test-time segments & -- & 96.2 / 97.3 & 96.7 \\
\bottomrule
\end{tabular}
\end{table}

\Cref{tab:sudoku-ablation} ablates the design choices of \ours on Sudoku-Extreme.
\begin{description}
  \item[Averaging.]
  The averaging is essential: without it, the single-state recursion overfits and loses $36$ points.
  \item[Spread of timescales.]
  A single shared rate recovers most of this, and the spread of timescales adds another $9$ points.
  The single rate $1/7$ is the geometric mean of $1$ and $\amin$.
  \item[Slow-only readout.]
  Reading only the slow channels adds $3$ points.
  Heads that read all channels overfit the training set more and vary more between seeds.
  \item[Rate jitter.]
  The rate jitter adds only little to the final accuracy.
  It does, however, reduce the late decline of the held-out accuracy: from its peak to the end of training it falls by $0.9$ points with jitter and by $2.3$ points without it (mean over two seeds).
  \item[Slowest rate.]
  The slowest rates $1/75$ and $1/100$ work as well as the reference $1/49$, and only the much faster $1/13$ is worse.
  \item[Test-time compute.]
  More segments at test time raise the accuracy to $96.7\%$ with $64$ segments.
\end{description}

\subsection{Training stability}
\label{sec:experiments:stability}


TRM trained with AdamW on Sudoku shows a late collapse: the training accuracy keeps increasing while the held-out accuracy drops.
None of the runs of \ours shows such a collapse (\cref{fig:convergence-baselines}).

\FloatBarrier
\section{Conclusion}
\label{sec:conclusion}

We have proposed STRM, a simpler and more stable variant of the Tiny Recursive Model.
It has shown strong performance on various puzzle-solving tasks under equal compute to other methods, with less overfitting than TRM and without its late decline of the held-out accuracy.
The improvements come from using a single state whose channels average the network outputs at distinct timescales and from reading out solutions from slow channels only.
We argue that fast channels serve as working memory and slow channels are responsible for compiling the answer.
We believe that \ours{} is a drop-in replacement for TRM and can be used in follow-up approaches, for example with winner-takes-all training~\citep{dillon2026speed}, parallel sampling~\citep{sghaier2026ptrm} and other improvements that are based on the original TRM architecture.

\newpage


{
    \small
    \bibliographystyle{ieeenat_fullname}
    \bibliography{references}

@misc{jolicoeur2025trm,
  title         = {Less is More: Recursive Reasoning with Tiny Networks},
  author        = {Jolicoeur-Martineau, Alexia},
  year          = {2025},
  howpublished  = {arXiv preprint arXiv:2510.04871},
  eprint        = {2510.04871},
  archiveprefix = {arXiv},
  primaryclass  = {cs.LG},
  url           = {https://arxiv.org/abs/2510.04871}
}

@misc{wang2025hrm,
  title         = {Hierarchical Reasoning Model},
  author        = {Wang, Guan and Li, Jin and Sun, Yuhao and Chen, Xing and Liu, Changling and Wu, Yue and Lu, Meng and Song, Sen and Abbasi Yadkori, Yasin},
  year          = {2025},
  howpublished  = {arXiv preprint arXiv:2506.21734},
  eprint        = {2506.21734},
  archiveprefix = {arXiv},
  primaryclass  = {cs.AI},
  url           = {https://arxiv.org/abs/2506.21734}
}

@misc{liao2026srm,
  title        = {Simple Recursive Model: Simplified, Single-State Reasoning with Skip Connections},
  author       = {Liao, Qianli and Poggio, Tomaso},
  year         = {2026},
  howpublished = {Research Square preprint},
  doi          = {10.21203/rs.3.rs-8492126/v1},
  url          = {https://doi.org/10.21203/rs.3.rs-8492126/v1},
  note         = {Code: \url{https://github.com/liaoq/SimpleRecursiveModel}}
}

@inproceedings{you2025paralleltts,
  title         = {Parallel Test-Time Scaling for Latent Reasoning Models},
  author        = {You, Runyang and Li, Yongqi and Liu, Meng and Wang, Wenjie and Nie, Liqiang and Li, Wenjie},
  booktitle     = {Annual Meeting of the Association for Computational Linguistics (ACL)},
  year          = {2026},
  eprint        = {2510.07745},
  archiveprefix = {arXiv},
  primaryclass  = {cs.CL},
  url           = {https://arxiv.org/abs/2510.07745}
}

@misc{gao2025urm,
  title         = {Universal Reasoning Model},
  author        = {Gao, Zitian and Chen, Lynx and Xiao, Yihao and Xing, He and Tao, Ran and Luo, Haoming and Zhou, Joey and Dai, Bryan},
  year          = {2025},
  howpublished  = {arXiv preprint arXiv:2512.14693},
  eprint        = {2512.14693},
  archiveprefix = {arXiv},
  primaryclass  = {cs.AI},
  url           = {https://arxiv.org/abs/2512.14693}
}

@misc{dillon2026speed,
  title         = {Speed is Confidence},
  author        = {Dillon, Joshua V.},
  year          = {2026},
  howpublished  = {arXiv preprint arXiv:2601.19085},
  eprint        = {2601.19085},
  archiveprefix = {arXiv},
  primaryclass  = {cs.LG},
  url           = {https://arxiv.org/abs/2601.19085}
}

@misc{wang2026trmmamba,
  title         = {Tiny Recursive Reasoning with {Mamba-2} Attention Hybrid},
  author        = {Wang, Wenlong and Reid, Fergal},
  year          = {2026},
  howpublished  = {arXiv preprint arXiv:2602.12078},
  eprint        = {2602.12078},
  archiveprefix = {arXiv},
  primaryclass  = {cs.AI},
  note          = {ICLR 2026 Workshop on Latent \& Implicit Thinking},
  url           = {https://arxiv.org/abs/2602.12078}
}

@misc{waugh2026ppbench,
  title         = {Pencil Puzzle Bench: A Benchmark for Multi-Step Verifiable Reasoning},
  author        = {Waugh, Justin},
  year          = {2026},
  howpublished  = {arXiv preprint arXiv:2603.02119},
  eprint        = {2603.02119},
  archiveprefix = {arXiv},
  primaryclass  = {cs.AI},
  url           = {https://arxiv.org/abs/2603.02119}
}

@misc{komisarczyk2026rim,
  title         = {Recursive Inference Machines for Neural Reasoning},
  author        = {Komisarczyk, Mieszko and Mathur, Saurabh and Kraus, Maurice and Natarajan, Sriraam and Kersting, Kristian},
  year          = {2026},
  howpublished  = {arXiv preprint arXiv:2603.05234},
  eprint        = {2603.05234},
  archiveprefix = {arXiv},
  primaryclass  = {cs.LG},
  url           = {https://arxiv.org/abs/2603.05234}
}

@misc{hakimi2026rsm,
  title         = {Form Follows Function: Recursive Stem Model},
  author        = {Hakimi, Navid},
  year          = {2026},
  howpublished  = {arXiv preprint arXiv:2603.15641},
  eprint        = {2603.15641},
  archiveprefix = {arXiv},
  primaryclass  = {cs.AI},
  url           = {https://arxiv.org/abs/2603.15641}
}

@inproceedings{efstathiou2026attractor,
  title     = {Recursive Reasoning as Attractor Landscape Search: Mechanistic Dynamics of the Tiny Recursive Model},
  author    = {Efstathiou, Andreas and Balwani, Aishwarya},
  booktitle = {ICLR 2026 Workshop on Latent \& Implicit Thinking -- Going Beyond CoT Reasoning},
  year      = {2026},
  url       = {https://openreview.net/forum?id=kKps9W1K7n}
}

@misc{sghaier2026ptrm,
  title         = {Probabilistic Tiny Recursive Model},
  author        = {Sghaier, Amin and Parviz, Ali and Jolicoeur-Martineau, Alexia},
  year          = {2026},
  howpublished  = {arXiv preprint arXiv:2605.19943},
  eprint        = {2605.19943},
  archiveprefix = {arXiv},
  primaryclass  = {cs.AI},
  url           = {https://arxiv.org/abs/2605.19943}
}

@misc{baek2026gram,
  title         = {Generative Recursive Reasoning},
  author        = {Baek, Junyeob and Jo, Mingyu and Kim, Minsu and Ren, Mengye and Bengio, Yoshua and Ahn, Sungjin},
  year          = {2026},
  howpublished  = {arXiv preprint arXiv:2605.19376},
  eprint        = {2605.19376},
  archiveprefix = {arXiv},
  primaryclass  = {cs.AI},
  url           = {https://arxiv.org/abs/2605.19376}
}

@inproceedings{huang2026eqr,
  title         = {Equilibrium Reasoners: Learning Attractors Enables Scalable Reasoning},
  author        = {Huang, Benhao and Geng, Zhengyang and Kolter, Zico},
  booktitle     = {International Conference on Machine Learning (ICML)},
  year          = {2026},
  eprint        = {2605.21488},
  archiveprefix = {arXiv},
  primaryclass  = {cs.LG},
  url           = {https://arxiv.org/abs/2605.21488}
}

@inproceedings{corbett2026guided,
  title         = {Boosting Inference with Guided Reasoning: Stochastic Exploration for Recursive Models},
  author        = {Corbett, Andrew and Sood, Archit and Tzatzopoulou, Anna and Ramesh, Sai-Aakash and Dodwell, Tim},
  year          = {2026},
  eprint        = {2605.25230},
  archiveprefix = {arXiv},
  primaryclass  = {cs.AI},
  url           = {https://arxiv.org/abs/2605.25230},
  booktitle     = {Advances in Neural Information Processing Systems},
  volume        = {40}
}

@misc{helbling2026frm,
  title         = {Flow Reasoning Models: Turning Flows Into Efficient Recurrent Reasoners},
  author        = {Helbling, Alec and Bryutkin, Andrey and Martino, Mauro and Chau, Duen Horng and Dehmamy, Nima and Strobelt, Hendrik},
  year          = {2026},
  howpublished  = {arXiv preprint arXiv:2606.29150},
  eprint        = {2606.29150},
  archiveprefix = {arXiv},
  primaryclass  = {cs.AI},
  url           = {https://arxiv.org/abs/2606.29150}
}

@misc{drozdova2026diffusion,
  title         = {Diffusion as a Training Curriculum for Timestep-Free Iterative Reasoning},
  author        = {Drozdova, Mariia and Sirbu, Aidan and Miotti, Pietro and Obryk, Robert and Etcheverry, Mayalen and Niklasson, Eyvind and Richards, Blake},
  year          = {2026},
  howpublished  = {arXiv preprint arXiv:2609.01449},
  note          = {NeurIPS 2026 Workshop on Beyond Next Token Prediction: Diffusion and Flow Models for Next-Generation Decoding},
  eprint        = {2609.01449},
  archiveprefix = {arXiv},
  primaryclass  = {cs.LG},
  url           = {https://arxiv.org/abs/2609.01449}
}

@misc{suleymanzade2026loopedflows,
  title         = {Thinking with Looped Flows},
  author        = {Suleymanzade, Ayhan and Lee, Chanhyuk and Eijkelboom, Floor and Boffi, Nicholas M. and Ceylan, {\.I}smail {\.I}lkan and Kim, Jinwoo},
  year          = {2026},
  howpublished  = {arXiv preprint arXiv:2609.11801},
  eprint        = {2609.11801},
  archiveprefix = {arXiv},
  primaryclass  = {cs.LG},
  url           = {https://arxiv.org/abs/2609.11801}
}

@inproceedings{dehghani2019ut,
  title     = {Universal Transformers},
  author    = {Dehghani, Mostafa and Gouws, Stephan and Vinyals, Oriol and Uszkoreit, Jakob and Kaiser, {\L}ukasz},
  booktitle = {International Conference on Learning Representations (ICLR)},
  year      = {2019},
  eprint    = {1807.03819},
  archiveprefix = {arXiv},
  url       = {https://arxiv.org/abs/1807.03819}
}

@misc{graves2016act,
  title         = {Adaptive Computation Time for Recurrent Neural Networks},
  author        = {Graves, Alex},
  year          = {2016},
  howpublished  = {arXiv preprint arXiv:1603.08983},
  eprint        = {1603.08983},
  archiveprefix = {arXiv},
  primaryclass  = {cs.NE},
  url           = {https://arxiv.org/abs/1603.08983}
}

@misc{banino2021pondernet,
  title         = {{PonderNet}: Learning to Ponder},
  author        = {Banino, Andrea and Balaguer, Jan and Blundell, Charles},
  year          = {2021},
  howpublished  = {arXiv preprint arXiv:2107.05407},
  eprint        = {2107.05407},
  archiveprefix = {arXiv},
  primaryclass  = {cs.LG},
  note          = {8th ICML Workshop on Automated Machine Learning (AutoML 2021)},
  url           = {https://arxiv.org/abs/2107.05407}
}

@inproceedings{bai2019deq,
  title     = {Deep Equilibrium Models},
  author    = {Bai, Shaojie and Kolter, J. Zico and Koltun, Vladlen},
  booktitle = {Advances in Neural Information Processing Systems},
  volume    = {32},
  year      = {2019},
  url       = {https://proceedings.neurips.cc/paper_files/paper/2019/hash/01386bd6d8e091c2ab4c7c7de644d37b-Abstract.html}
}

@inproceedings{koutnik2014clockwork,
  title     = {A Clockwork {RNN}},
  author    = {Koutn{\'\i}k, Jan and Greff, Klaus and Gomez, Faustino and Schmidhuber, J{\"u}rgen},
  booktitle = {Proceedings of the 31st International Conference on Machine Learning},
  series    = {Proceedings of Machine Learning Research},
  volume    = {32},
  pages     = {1863--1871},
  publisher = {PMLR},
  year      = {2014},
  url       = {https://proceedings.mlr.press/v32/koutnik14.html}
}

@inproceedings{chung2017hmrnn,
  title     = {Hierarchical Multiscale Recurrent Neural Networks},
  author    = {Chung, Junyoung and Ahn, Sungjin and Bengio, Yoshua},
  booktitle = {International Conference on Learning Representations (ICLR)},
  year      = {2017},
  eprint    = {1609.01704},
  archiveprefix = {arXiv},
  url       = {https://arxiv.org/abs/1609.01704}
}

@article{jaeger2007leaky,
  title   = {Optimization and Applications of Echo State Networks with Leaky-Integrator Neurons},
  author  = {Jaeger, Herbert and Luko{\v{s}}evi{\v{c}}ius, Mantas and Popovici, Dan and Siewert, Udo},
  journal = {Neural Networks},
  volume  = {20},
  number  = {3},
  pages   = {335--352},
  year    = {2007},
  doi     = {10.1016/j.neunet.2007.04.016}
}

@article{polyak1992averaging,
  title   = {Acceleration of Stochastic Approximation by Averaging},
  author  = {Polyak, Boris T. and Juditsky, Anatoli B.},
  journal = {SIAM Journal on Control and Optimization},
  volume  = {30},
  number  = {4},
  pages   = {838--855},
  year    = {1992},
  doi     = {10.1137/0330046}
}

@article{su2024roformer,
  title   = {{RoFormer}: Enhanced Transformer with Rotary Position Embedding},
  author  = {Su, Jianlin and Lu, Yu and Pan, Shengfeng and Murtadha, Ahmed and Wen, Bo and Liu, Yunfeng},
  journal = {Neurocomputing},
  volume  = {568},
  pages   = {127063},
  year    = {2024},
  doi     = {10.1016/j.neucom.2023.127063},
  eprint  = {2104.09864},
  archiveprefix = {arXiv}
}

@misc{shazeer2020glu,
  title         = {{GLU} Variants Improve Transformer},
  author        = {Shazeer, Noam},
  year          = {2020},
  howpublished  = {arXiv preprint arXiv:2002.05202},
  eprint        = {2002.05202},
  archiveprefix = {arXiv},
  primaryclass  = {cs.LG},
  url           = {https://arxiv.org/abs/2002.05202}
}

@inproceedings{zhang2019rmsnorm,
  title     = {Root Mean Square Layer Normalization},
  author    = {Zhang, Biao and Sennrich, Rico},
  booktitle = {Advances in Neural Information Processing Systems},
  volume    = {32},
  year      = {2019},
  url       = {https://proceedings.neurips.cc/paper_files/paper/2019/hash/1e8a19426224ca89e83cef47f1e7f53b-Abstract.html}
}

@inproceedings{loshchilov2019adamw,
  title     = {Decoupled Weight Decay Regularization},
  author    = {Loshchilov, Ilya and Hutter, Frank},
  booktitle = {International Conference on Learning Representations (ICLR)},
  year      = {2019},
  eprint    = {1711.05101},
  archiveprefix = {arXiv},
  url       = {https://arxiv.org/abs/1711.05101}
}

@misc{ge2025hrmmisconceptions,
  title         = {Hierarchical Reasoning Models: Perspectives and Misconceptions},
  author        = {Ge, Renee and Liao, Qianli and Poggio, Tomaso},
  year          = {2025},
  howpublished  = {arXiv preprint arXiv:2510.00355},
  eprint        = {2510.00355},
  archiveprefix = {arXiv},
  primaryclass  = {cs.AI},
  url           = {https://arxiv.org/abs/2510.00355}
}

@misc{arcprize2025hrmanalysis,
  title        = {The Hidden Drivers of {HRM}'s Performance on {ARC-AGI}},
  author       = {{ARC Prize Team}},
  year         = {2025},
  howpublished = {ARC Prize blog, 15 August 2025},
  url          = {https://arcprize.org/blog/hrm-analysis}
}

@misc{qasim2025cgar,
  title         = {Accelerating Training Speed of Tiny Recursive Models with Curriculum Guided Adaptive Recursion},
  author        = {Qasim, Kaleem Ullah and Zhang, Jiashu},
  year          = {2025},
  howpublished  = {arXiv preprint arXiv:2511.08653},
  eprint        = {2511.08653},
  archiveprefix = {arXiv},
  primaryclass  = {cs.LG},
  url           = {https://arxiv.org/abs/2511.08653}
}

@inproceedings{asadulaev2025policy,
  title         = {Latent Reasoning in {TRMs} is Secretly a Policy Improvement Operator},
  author        = {Asadulaev, Arip and Banerjee, Rayan and Karray, Fakhri and Takac, Martin},
  booktitle     = {International Conference on Machine Learning (ICML)},
  year          = {2026},
  eprint        = {2511.16886},
  archiveprefix = {arXiv},
  primaryclass  = {cs.CL},
  url           = {https://arxiv.org/abs/2511.16886}
}

@misc{ren2026guessing,
  title         = {Are Your Reasoning Models Reasoning or Guessing? {A} Mechanistic Analysis of Hierarchical Reasoning Models},
  author        = {Ren, Zirui and Liu, Ziming},
  year          = {2026},
  howpublished  = {arXiv preprint arXiv:2601.10679},
  eprint        = {2601.10679},
  archiveprefix = {arXiv},
  primaryclass  = {cs.AI},
  url           = {https://arxiv.org/abs/2601.10679}
}

@article{eskin2026dynamical,
  title   = {The Contraction Mapping Model: Ultra-Compact Algorithmic Reasoning via Neural Stochastic Differential Equations},
  author  = {Es'kin, Vasiliy A. and Smorkalov, Mikhail E.},
  journal = {Neurocomputing},
  volume  = {709},
  pages   = {135271},
  year    = {2027},
  doi     = {10.1016/j.neucom.2026.135271},
  eprint  = {2603.22871},
  archiveprefix = {arXiv}
}

@inproceedings{prairie2026parcae,
  title         = {Parcae: Scaling Laws For Stable Looped Language Models},
  author        = {Prairie, Hayden and Novack, Zachary and Berg-Kirkpatrick, Taylor and Fu, Daniel Y.},
  booktitle     = {Conference on Language Modeling (COLM)},
  year          = {2026},
  eprint        = {2604.12946},
  archiveprefix = {arXiv},
  primaryclass  = {cs.LG},
  url           = {https://arxiv.org/abs/2604.12946}
}

@misc{han2026hierflat,
  title         = {Hierarchical vs. Flat Iteration in Shared-Weight Transformers},
  author        = {Han, Sang-Il},
  year          = {2026},
  howpublished  = {arXiv preprint arXiv:2604.14442},
  eprint        = {2604.14442},
  archiveprefix = {arXiv},
  primaryclass  = {cs.CL},
  url           = {https://arxiv.org/abs/2604.14442}
}

@misc{sapunov2026utmemory,
  title         = {Universal Transformers Need Memory: Depth-State Trade-offs in Adaptive Recursive Reasoning},
  author        = {Sapunov, Grigory},
  year          = {2026},
  howpublished  = {arXiv preprint arXiv:2604.21999},
  eprint        = {2604.21999},
  archiveprefix = {arXiv},
  primaryclass  = {cs.LG},
  url           = {https://arxiv.org/abs/2604.21999}
}

@misc{buehlmaier2026layernorm,
  title         = {Spectral Margins and Memory Routing in Looped Transformers},
  author        = {Buehlmaier, Matthias M. M.},
  year          = {2026},
  howpublished  = {arXiv preprint arXiv:2607.10681},
  eprint        = {2607.10681},
  archiveprefix = {arXiv},
  primaryclass  = {cs.LG},
  url           = {https://arxiv.org/abs/2607.10681}
}

@misc{rodrigues2026dissecting,
  title         = {Dissecting Hierarchical Reasoning Models: A Mechanistic Study},
  author        = {Rodrigues, Leo Raphael and Kang, Jian},
  year          = {2026},
  howpublished  = {arXiv preprint arXiv:2609.22197},
  note          = {ICML 2026 Workshop on Mechanistic Interpretability},
  eprint        = {2609.22197},
  archiveprefix = {arXiv},
  primaryclass  = {cs.LG},
  url           = {https://arxiv.org/abs/2609.22197}
}

@misc{krishna2026finish,
  title         = {When Recursive Models Finish Computing},
  author        = {Krishna, Hare and Singh, Shubham and Ebert, Stephen and Sun, Hao-Yu},
  year          = {2026},
  howpublished  = {arXiv preprint arXiv:2609.26487},
  eprint        = {2609.26487},
  archiveprefix = {arXiv},
  primaryclass  = {cs.LG},
  url           = {https://arxiv.org/abs/2609.26487}
}

@misc{shakhvalieva2026limits,
  title         = {What Limits Recursive Reasoning Models: Optimization, Architecture and Test-Time Scaling},
  author        = {Shakhvalieva, Yuliana and Kharchev, Dmitrii and Bezrukov, Viacheslav and Fedorova, Inessa and Bocharov, Dmitry and Oseledets, Ivan and Ternovskii, Valerii},
  year          = {2026},
  howpublished  = {arXiv preprint arXiv:2609.39967},
  eprint        = {2609.39967},
  archiveprefix = {arXiv},
  primaryclass  = {cs.LG},
  url           = {https://arxiv.org/abs/2609.39967}
}

@misc{movahedi2026fprm,
  title         = {Fixed-Point Reasoners: Stable and Adaptive Deep Looped Transformers},
  author        = {Movahedi, Sajad and Milovanovi{\'c}, Vera and Feigin, Shlomo Libo and Theus, Alexander and Hofmann, Thomas and Boeva, Valentina and Rusch, T. Konstantin and Orvieto, Antonio},
  year          = {2026},
  howpublished  = {arXiv preprint arXiv:2606.18206},
  eprint        = {2606.18206},
  archiveprefix = {arXiv},
  url           = {https://arxiv.org/abs/2606.18206}
}

@misc{zela2026sfprm,
  title         = {Spherical Fixed-Point Reasoners: Adaptive Loop Halting on the Hypersphere},
  author        = {Zela, Arber and Maranjyan, Artavazd and Puigdemont, Pol and Berghaus, David and Boreiko, Valentyn and Cevher, Volkan},
  year          = {2026},
  howpublished  = {arXiv preprint arXiv:2610.25646},
  eprint        = {2610.25646},
  archiveprefix = {arXiv},
  url           = {https://arxiv.org/abs/2610.25646}
}

@misc{chollet2019measure,
  title         = {On the Measure of Intelligence},
  author        = {Chollet, Fran{\c{c}}ois},
  year          = {2019},
  howpublished  = {arXiv preprint arXiv:1911.01547},
  eprint        = {1911.01547},
  archiveprefix = {arXiv},
  url           = {https://arxiv.org/abs/1911.01547}
}

@misc{chollet2025arcagi2,
  title         = {{ARC-AGI-2}: A New Challenge for Frontier {AI} Reasoning Systems},
  author        = {Chollet, Fran{\c{c}}ois and Knoop, Mike and Kamradt, Gregory and Landers, Bryan and Pinkard, Henry},
  year          = {2025},
  howpublished  = {arXiv preprint arXiv:2505.11831},
  eprint        = {2505.11831},
  archiveprefix = {arXiv},
  url           = {https://arxiv.org/abs/2505.11831}
}

@article{moskvichev2023conceptarc,
  title         = {The {ConceptARC} Benchmark: Evaluating Understanding and Generalization in the {ARC} Domain},
  author        = {Moskvichev, Arseny and Odouard, Victor Vikram and Mitchell, Melanie},
  year          = {2023},
  eprint        = {2305.07141},
  archiveprefix = {arXiv},
  url           = {https://arxiv.org/abs/2305.07141},
  journal       = {Transactions on Machine Learning Research},
}
}

\end{document}